\documentclass[conference]{IEEEtran}
\usepackage{colortbl}
\usepackage{xcolor}

\usepackage{cite}

\ifCLASSINFOpdf
  \usepackage[pdftex]{graphicx}
\else
  \usepackage[dvips]{graphicx}
\fi
\usepackage{amsmath}
\usepackage{algorithmic}
\ifCLASSOPTIONcompsoc
  \usepackage[caption=false,font=normalsize,labelfont=sf,textfont=sf]{subfig}
\else
  \usepackage[caption=false,font=footnotesize]{subfig}
\fi
\usepackage{url}

\begin{document}
%
\title{Deep RTS: A Game Environment for\\Deep Reinforcement Learning in\\Real-Time Strategy Games}

\author{\IEEEauthorblockN{Per-Arne Andersen}
\IEEEauthorblockA{\textit{Department of ICT} \\
	\textit{University of Agder}\\
	Grimstad, Norway \\
	per.andersen@uia.no}
\and

\IEEEauthorblockN{Morten Goodwin}
\IEEEauthorblockA{\textit{Department of ICT} \\
	\textit{University of Agder}\\
	Grimstad, Norway \\
	morten.goodwin@uia.no}
\and

\IEEEauthorblockN{Ole-Christoffer Granmo}
\IEEEauthorblockA{\textit{Department of ICT} \\
	\textit{University of Agder}\\
	Grimstad, Norway \\
	ole.granmo@uia.no}

}

\maketitle

\begin{abstract}
Reinforcement learning (RL) is an area of research that has blossomed tremendously in recent years and has shown remarkable potential for artificial intelligence based opponents in computer games. This success is primarily due to the vast capabilities of convolutional neural networks, that can extract useful features from noisy and complex data. Games are excellent tools to test and push the boundaries of novel RL algorithms because they give valuable insight into how well an algorithm can perform in isolated environments without the real-life consequences. Real-time strategy games (RTS) is a genre that has tremendous complexity and challenges the player in short and long-term planning. There is much research that focuses on applied RL in RTS games, and novel advances are therefore anticipated in the not too distant future. However, there are to date few environments for testing RTS AIs. Environments in the literature are often either overly simplistic, such as microRTS, or complex and without the possibility for accelerated learning on consumer hardware like StarCraft II. This paper introduces the Deep RTS game environment for testing cutting-edge artificial intelligence algorithms for RTS games. Deep RTS is a high-performance RTS game made specifically for artificial intelligence research. It supports accelerated learning, meaning that it can learn at a magnitude of 50 000 times faster compared to existing RTS games. Deep RTS has a flexible configuration, enabling research in several different RTS scenarios, including partially observable state-spaces and map complexity. We show that Deep RTS lives up to our promises by comparing its performance with microRTS, ELF, and StarCraft II on high-end consumer hardware. Using Deep RTS, we show that a Deep Q-Network agent beats random-play agents over 70\% of the time. Deep RTS is publicly available at \url{https://github.com/cair/DeepRTS}.

\end{abstract}


%
\IEEEpeerreviewmaketitle

\section{Introduction}
\label{sec:introduction}
Despite many advances in Artificial Intelligence (AI) for games, no universal Reinforcement learning (RL) algorithm can be applied to complex game environments without extensive data manipulation or customization. This includes traditional Real-time strategy games (RTS) such as WarCraft III, StarCraft II, and Age of Empires. RL has recently been applied to simpler game environments such as those found in the Arcade Learning Environment~\cite{Mnih2013}(ALE) and board games~\cite{Silver2017} but has not successfully been applied to more advanced games. Further, existing game environments that target AI research are either overly simplistic such as ALE or complex such as StarCraft II.

RL has in recent years had tremendous progress in learning how to control agents from high-dimensional sensory inputs like images. In simple environments, this has been proven to work well~\cite{Mnih2015}, but are still an issue for complex environments with large state and action spaces~\cite{Mirowski2016}. The distinction between simple and complex tasks in RL often lies in how easy it is to design a reward model that encourages the algorithm to improve its policy without ending in local optima~\cite{Kaelbling1996}. For simple tasks, the reward function can be described by only a few parameters, while in more demanding tasks, the algorithm struggles to determine what the reward signal is trying to accomplish~\cite{Konidaris2006}. For this reason, the reward function is in literature often a constant or single-valued variable for most time-steps, where only the final time-step determines a negative or positive reward~\cite{Tesauro1994, Vinyals2017, Silver2016}. In this paper we introduce Deep RTS, a new game environment targeted deep reinforcement learning (DRL) research. Deep RTS is an RTS simulator inspired by the famous StarCraft II video game by Blizzard Entertainment.

This paper is structured as follows. First, Section~\ref{sec:related_game_environments} and Section~\ref{sec:reinforcement_learning_in_games} thoroughly outlines previous work and central achievements using game environments for RL research. Next, Section~\ref{sec:deep_rts} introduces the Deep RTS game environment. Section~\ref{sec:experiments} presents the Deep RTS performance, a comparison between well-established game environments and Deep RTS, and experimental results using Deep Q-Network as an agent in Deep RTS. Subsequently, Section~\ref{sec:conclusion} concludes the contribution of this paper and outlines a roadmap for future work.


\section{Related Game Environments}
\label{sec:related_game_environments}
There exist several exciting game environments in the literature that focus on state-of-the-art research in AI algorithms. Few game environments target the RTS-genre. One the reason may be because these environments are by nature challenging to solve, and there are few ways to fit results with preprocessing tricks. It is, however, essential to include RTS as part of the active research of deep reinforcement learning algorithms as they feature long-term planning. This section outlines a thorough literature review of existing game platforms and environments and is summarized in Table~\ref{tbl:game_environments}.

\begin{table}[!ht]
	\centering
	\caption{Selected game environments that is actively used in reinforcement learning research}
	\label{tbl:game_environments}
	\begin{tabular}{|cccccc|}
		\hline
		\rowcolor{black}
		\color{white}\textbf{Platform} & \color{white}\textbf{RTS} & \color{white}\textbf{Complex}\footnotemark & \color{white}\textbf{Year} & \color{white}\textbf{Solved} & \color{white}\textbf{Source} \\ \hline

		\multicolumn{1}{|l|}{ALE} & No & No & 2012 & Yes & \cite{Bellemare2012} \\
		\multicolumn{1}{|l|}{Malmo Platform} & No & Yes & 2016 & No & \cite{Johnson2016} \\
		\multicolumn{1}{|l|}{ViZDoom} & No & Yes & 2016 & No & \cite{Kempka2016a}\ \\
		\multicolumn{1}{|l|}{DeepMind Lab} & No & Yes & 2016 & No & \cite{Beattie2016} \\
		\multicolumn{1}{|l|}{OpenAI Gym} & No & No & 2016 & No & \cite{Brockman2016} \\
		\multicolumn{1}{|l|}{OpenAI Universe} & No & Yes & 2016 & No & \cite{OpenAI2016} \\
		\hline
		\multicolumn{1}{|l|}{Stratagus} & Yes & Yes & 2005 & No & \cite{Ponsen2005 } \\
		\multicolumn{1}{|l|}{microRTS} & Yes & No & 2013 & No & \cite{Ontanon2013}\\
		\multicolumn{1}{|l|}{TorchCraft} & Yes & Yes & 2016 & No & \cite{Synnaeve2016} \\
		\multicolumn{1}{|l|}{ELF} & Yes & Yes & 2017 & No & \cite{Tian2017} \\
		\multicolumn{1}{|l|}{SC2LE} & Yes & Yes & 2017 & No & \cite{Vinyals2017} \\
		\multicolumn{1}{|l|}{\textbf{Deep RTS}} & Yes & Yes & 2018 & No & - \\
		\hline
	\end{tabular}
\end{table}
\footnotetext{A Complex environment has an enormous state-space, with reward signals that are difficult to correlate to an action.}

\subsection{Stratagus}
\label{subsec:rge:stratagus}
Stratagus is an open source game engine that can be used to create RTS-themed games. Wargus, a clone of Warcraft II, and Stargus, a clone of StarCraft I are examples of games implemented in the Stratagus game engine. Stratagus is not an engine that targets machine learning explicitly, but several researchers have performed experiments in case-based reasoning \cite{Ontanon2008, Fathy2010} and q-learning \cite{Jaidee2012} using Wargus. Stratagus is still actively updated by contributions from the community.

\subsection{Arcade Learning Environment}
Bellemare \textit{et al}. provided in 2012 the arcade learning environment that enabled researchers to conduct cutting-edge research in general deep learning \cite{Bellemare2012}. The package provided hundreds of Atari 2600 environments that in 2013 allowed Minh \textit{et al}. to do a breakthrough using Deep Q-Learning and A3C. The platform has been a critical component in several advances in RL research. \cite{Mnih2013, Mnih2015, Mnih2016}

\subsection{microRTS}
microRTS is a simple RTS game, designed to conduct AI research. The idea behind microRTS is to strip away the computational heavy game logic to increase the performance and to enable researchers to test theoretical concepts quickly \cite{Ontanon2013}. The microRTS game logic is deterministic, and include options for full and partially-observable state-spaces. The primary field of research in microRTS is game-tree search techniques such as variations of Monte-Carlo tree search and minimax \cite{Barriga2017, Ontanon2013, Shleyfman2014}.

\subsection{TorchCraft}
In 2016, a research group developed TorchCraft, a bridge that enables research in the game StarCraft. TorchCraft intends to provide the reinforcement learning community with a way to allow research on complex systems where only a fraction of the state-space is available \cite{Synnaeve2016}. In literature, TorchCraft has been used for deep learning research \cite{Churchill2017, Peng2017}. There is also a dataset that provides data from over 65,000 StarCraft replays \cite{Lin2017}.

\subsection{Malmo Platform}
The Malmo project is a platform built atop of the popular game \textit{Minecraft}. This game is set in a 3D environment where the object is to survive in a world of dangers. The paper \textit{The Malmo Platform for Artificial Intelligence Experimentation} by Johnson \textit{et al}. claims that the platform has all characteristics qualifying it to be a platform for general artificial intelligence research.\cite{Johnson2016}

\subsection{ViZDoom}
ViZDoom is a platform for research in visual reinforcement learning. With the paper \textit{ViZDoom: A Doom-based AI Research Platform for Visual Reinforcement Learning} Kempka \textit{et al}. illustrated that an RL agent could successfully learn to play the game \textit{Doom}, a first-person shooter game, with behavior similar to humans.~\cite{Kempka2016}

\subsection{DeepMind Lab}
With the paper \textit{DeepMind Lab}, Beattie \textit{et al}. released a platform for 3D navigation and puzzle solving tasks. The primary purpose of DeepMind Lab is to act as a platform for DRL research.\cite{Beattie2016}

\subsection{OpenAI Gym}
In 2016, Brockman \textit{et al}. from OpenAI released GYM which they referred to as \textit{"a toolkit for developing and comparing reinforcement learning algorithms"}. GYM provides various types of environments from following technologies: Algorithmic tasks, Atari 2600, Board games, Box2d physics engine, MuJoCo physics engine, and Text-based environments.
OpenAI also hosts a website where researchers can submit their performance for comparison between algorithms. GYM is open-source and encourages researchers to add support for their environments. \cite{Brockman2016}

\subsection{OpenAI Universe}
OpenAI recently released a new learning platform called \textit{Universe}. This environment further adds support for environments running inside VNC. It also supports running Flash games and browser applications. However, despite OpenAI's open-source policy, they do not allow researchers to add new environments to the repository. This limits the possibilities of running any environment. The OpenAI Universe is, however, a significant learning platform as it also has support for desktop games like Grand Theft Auto IV, which allow for research in autonomous driving \cite{Li2017}.

\subsection{ELF}
The Extensive Lightweight Flexible (ELF) research platform was recently present at NIPS with the paper \textit{ELF: An Extensive, Lightweight and Flexible Research Platform for Real-time Strategy Games}. This paper focuses on RTS game research and is the first platform officially targeting these types of games. \cite{Tian2017}

\subsection{StarCraft II Learning Environment}
SC2LE (StarCraft II Learning Environment) is an API wrapper that facilitates access to the StarCraft II game-state using languages such as Python. The purpose is to enable reinforcement learning and machine learning algorithms to be used as AI for the game players. StarCraft II is a complex environment that requires short and long-term planning. It is difficult to observe a correlation between actions and rewards due to the imperfect state information and delayed rewards, making StarCraft II one of the hardest challenges so far in AI research.


\section{Reinforcement Learning in Games}
\label{sec:reinforcement_learning_in_games}
Although there are several open-source game environments suited for reinforcement learning, few of them are part of a success story. One of the reasons for this is that current state-of-the-art algorithms are seemingly unstable \cite{Li2017}, and have difficulties to converge towards optimal policy in environments with multi-reward objectives \cite{Garcia2015}. This section exhibits the most significant achievements using reinforcement learning in games.

\subsection{TD-Gammon}
TD-Gammon is an algorithm capable of reaching an expert level of play in the board game \textit{Backgammon}~\cite{Tesauro1994, Tesauro1995}. The algorithm was developed by Gerald Tesauro in 1992 at IBM's Thomas J. Watson Research Center. TD-Gammon consists of a three-layer artificial neural network (ANN) and is trained using a reinforcement learning technique called \textit{TD-Lambda}. TD-Lambda is a temporal difference learning algorithm invented by Richard S. Sutton~\cite{Sutton1990}. The ANN iterates over all possible moves the player can perform and estimates the reward for that particular move. The action that yields the highest reward is then selected. TD-Gammon is the first algorithm to utilize self-play methods to improve the ANN parameters.

\subsection{AlphaGO}
In late 2015, \textit{AlphaGO} became the first algorithm to win against a human professional Go player. AlphaGO is a reinforcement learning framework that uses Monte-Carlo tree search and two deep neural networks for value and policy estimation~\cite{Silver2016}. Value refers to the expected future reward from a state assuming that the agent plays perfectly. The policy network attempts to learn which action is best in any given board configuration. The earliest versions of AlphaGO used training data from previous games played by human professionals. In the most recent version, \textit{AlphaGO Zero}, only self-play is used to train the AI~\cite{Silver2017a}. In a recent update, AlphaGO was generalized to work for Chess and Shogi (Japanese Chess) only using 24 hours to reach a superhuman level of play~\cite{Silver2017}.

\subsection{DeepStack}
DeepStack is an algorithm that can perform an expert level play in Texas Hold'em poker. This algorithm uses tree-search in conjunction with neural networks to perform sensible actions in the game \cite{Moravcik2017a}. DeepStack is a general-purpose algorithm that aims to solve problems with imperfect information. The DeepStack algorithm is open-source and available at \url{https://github.com/lifrordi/DeepStack-Leduc}.

\subsection{Dota 2}
DOTA 2 is a complex player versus player game where the player controls a hero unit. The game objective is to defeat the enemy heroes and destroy their base. In August 2017, OpenAI invented a reinforcement learning based AI that defeated professional players in one versus one games. The training was done by only using self-play, and the algorithm learned how to exploit game mechanics to perform well within the environment. DOTA 2 is used actively in research where the next goal is to train the AI to play in a team-game based environment.

\section{The Deep RTS Learning Environment}
\label{sec:deep_rts}
There is a need for new RTS game environments targeting reinforcement learning research. Few game environments have a complexity suited for current state-of-the-art research, and there is a lack of flexibility the existing solutions.

The Deep RTS game environment enables research at different difficulty levels in planning, reasoning, and control. The inspiration behind this contribution is microRTS and StarCraft II, where the goal is to create an environment that features challenges between the two. The simplest configurations of Deep RTS are deterministic and non-durative. Actions in the non-durative configuration are directly applied to the environment within the next few game frames. This makes the correlation between action and reward easier to observe. The durative configuration complicates the state-space significantly because it then becomes a temporal problem that requires long-term planning. Deep RTS supports the OpenAI Gym abstraction through the Python API and is a promising tool for reinforcement learning research.

\subsection{Game Objective}
The objective of the Deep RTS challenge is to build a base consisting of a town-hall, and then strive to expand the base using gathered resources to gain the military upper hand. Military units are used to conduct attacks where the primary goal is to demolish the base of the opponent. Players start with a worker unit. The primary objective of the worker units is to expand the base offensive, defensive and to gather natural resources found throughout the game world. Buildings can further spawn additional units that strengthen the offensive capabilities of the player.  For a player to reach the terminal state, all opponent units must be destroyed.

A regular RTS game can be represented in three stages: early-game, mid-game and late-game. Early-game is the gathering and base expansion stage. The mid-game focuses on the military and economic superiority, while the late-game stage is usually a deathmatch between the players until the game ends.

\begin{table}[!ht]
	\centering
	
	\caption{An overview of available scenarios found in the Deep RTS game environment}
	\label{tbl:scenario_overview}
	\begin{tabular}{|c|c|c|c|}
		\hline
		\rowcolor{black}
		\color{white}Scenario Name & \color{white}Description & \color{white}Game Length & \color{white}Map Size \\
		\hline
		10x10-2-FFA& 2-Player game & 600-900 ticks & 10x10 \\
		\hline
		15x15-2-FFA& 2-Player game & 900-1300 ticks & 15x15 \\
		\hline
		21x21-2-FFA& 2-Player game & 2000-3000 ticks & 21x21 \\
		\hline
		31x31-2-FFA& 2-Player game & 6000-9000 ticks & 31x31 \\
		\hline
		31x31-4-FFA& 4-Player game & 8000-11k ticks & 31x31 \\
		\hline
		31x31-6-FFA& 6-Player game & 15k-20k ticks & 31x31 \\
		\hline
		solo-score & Score Accumulation  & 1200 ticks & 10x10 \\
		\hline
		solo-resources & Resource Harvesting  & 600 ticks & 10x10 \\
		\hline
		solo-army & Army Accumulation  & 1200 ticks & 10x10 \\
		\hline
	\end{tabular}
	
\end{table}

Because Deep RTS targets a various range of reinforcement learning tasks, there are game scenarios such as resource gathering tasks, military tasks, and defensive tasks that narrows the complexity of a full RTS game. Table \ref{tbl:scenario_overview} shows nine scenarios currently implemented in the Deep RTS game environment. The first six scenarios are regular RTS games with the possibility of having 6 active players in a free-for-all setting. The \textit{solo-score} scenario features an environment where the objective is to only generate as much score as possible in shortest amount of time. \textit{solo-resources} is a game mode that focuses on resource gathering. The agent must find a balance between base expansion and resource gathering to optimally gather as many resources as possible. \textit{solo-army} is a scenario where the primary goal is to expand the military forces quickly and launch an attack on an idle enemy. The Deep RTS game environment enables researchers to create custom scenarios via a flexible configuration interface.

\subsection{Game Mechanics}
\begin{table}[!ht]
	\centering
	
	\caption{Central configuration flags for the Deep RTS game engine}
	\label{tbl:game_mechanic_conf}
	\begin{tabular}{|c|c|c|}
		\hline
		\rowcolor{black}
		\color{white}Config Name & \color{white}Type & \color{white}Description \\
		\hline
		instant\_town\_hall & Bool & Spawn Town-Hall at game start.\\
		\hline
		instant\_building & Bool & Non-durative Build Mode.\\
		\hline
		instant\_walking & Bool & Non-durative Walk Mode.\\
		\hline
		harvest\_forever & Bool & Harvest resources automatically.\\
		\hline
		auto\_attack & Bool & Automatic retaliation when being attacked.\\
		\hline
		durative & Bool & Enable durative mode.\\
		\hline
	\end{tabular}
	
\end{table}

The game mechanics of the Deep RTS are flexible and can be adjusted before a game starts. Table \ref{tbl:game_mechanic_conf} shows a list of configurations currently available. An important design choice is to allow actions to affect the environment without any temporal delay. All actions are bound to a tick-timer that defaults to 10, that is, it takes 10 ticks for a unit to move one tile, 10 ticks for a unit to attack once, and 300 ticks to build buildings. The tick-timer also includes a multiplier that enables adjustments of how many ticks equals a second. For each iteration of the game-loop, the tick counter is incremented, and the tick-timers are evaluated. By using tick-timers, the game-state resembles how the StarCraft II game mechanics function while lowering the tick-timer value better resembles microRTS.

\begin{figure}[!ht]
	\centering
	\includegraphics[width=\linewidth]{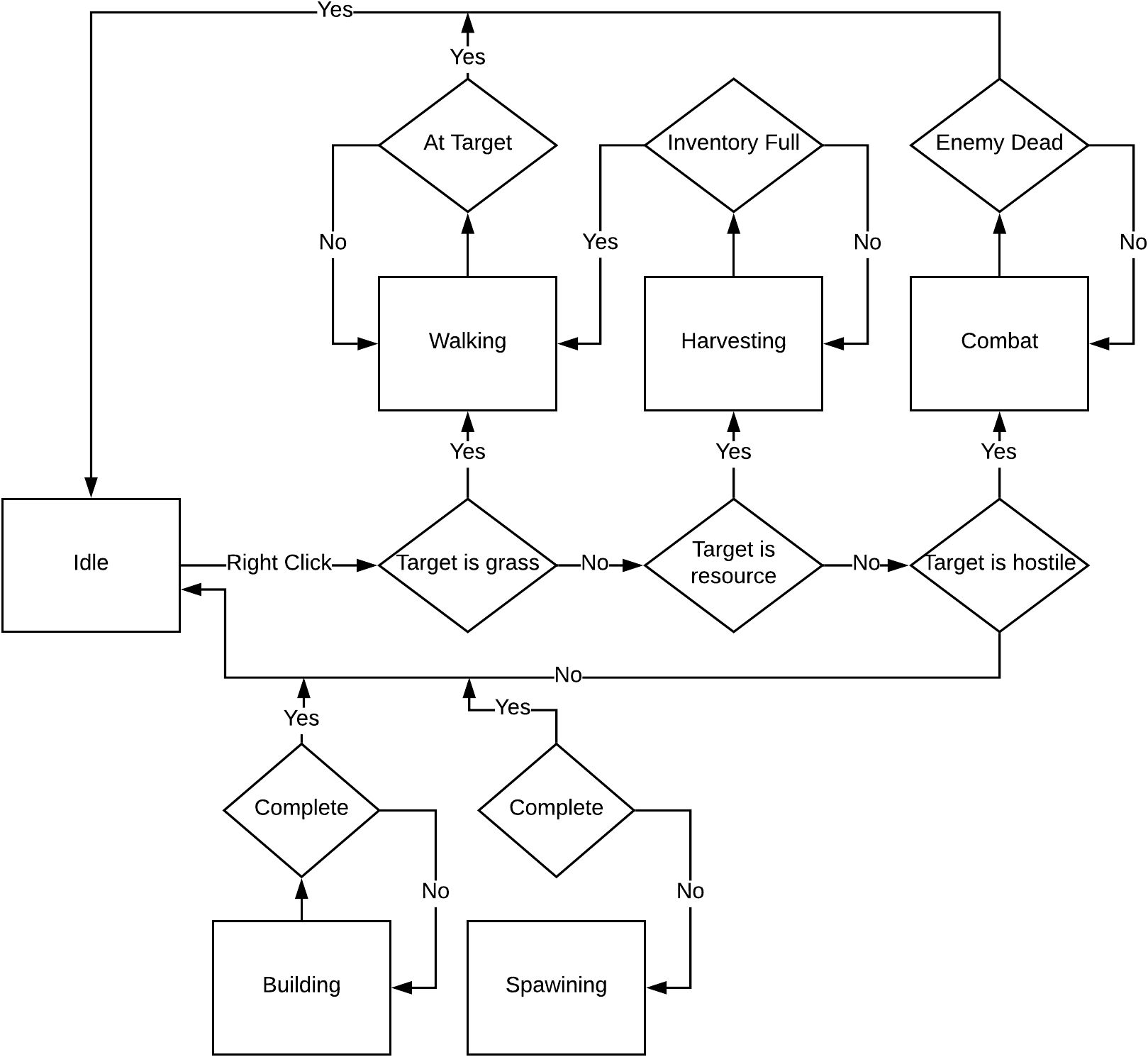}
	\caption{Unit state evaluation based on actions and current state}
	\label{fig:deeprts_mechanics_flowchart}
\end{figure}

All game entities (Units and Buildings) have a state-machine that determine its current state. Figure \ref{fig:deeprts_mechanics_flowchart} illustrates a portion of the logic that is evaluated through the state-machine. Entities start in the Spawning state transitioning to the Idle state when the entity spawn process is complete. The Idle state can be considered the default state of all entities and is only transitioned from when the player interacts with the entity. This implementation enables researchers to modify the state-transitions to produce alternative game logic.

\begin{table}[!ht]
	\centering
	\caption{The available economic resources and limits available to players in Deep RTS}
	\label{tbl:deeprts_resources}
	\begin{tabular}{|l|l|l|l|l|l|l}
		\hline
		\rowcolor{black}
		\multicolumn{6}{|c|}{{\color{white}Player Resources}} \\
		\hline
		\textbf{Property} & \textbf{Lumber} & \textbf{Gold} & \textbf{Oil} & \textbf{Food} & \textbf{Units} \\
		\hline
		\textbf{Range} & 0 - $10^6$ & 0 - $10^6$ & 0 - $10^6$ & 0 - 6000 & 0 - 2000 \\
		\hline
	\end{tabular}
	
\end{table}

Table \ref{tbl:deeprts_resources} shows the available resources and unit limits in the Deep RTS game environment. There are primarily three resources, gold, lumber, and oil that are available for workers to harvest. The value range is practically limited to the number of resources that exist on the game map. The food limit and the unit limit ensures that the player does not produce units excessively.

\subsection{Graphics}
\begin{figure}[!t]
	\centering
	\includegraphics[width=0.9\linewidth]{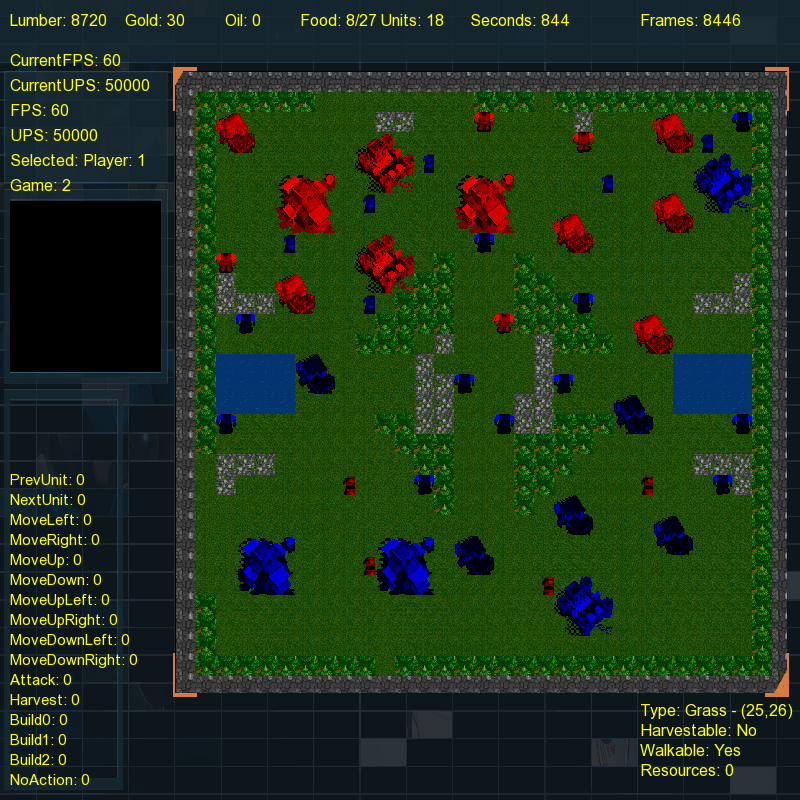}
	\caption{Overview of a battle in the fully-observable Deep RTS state-space using the C++ graphical user interface}
	\label{fig:deep_rts_game}
\end{figure}

\begin{figure}[!t]
	\centering
	\includegraphics[width=\linewidth]{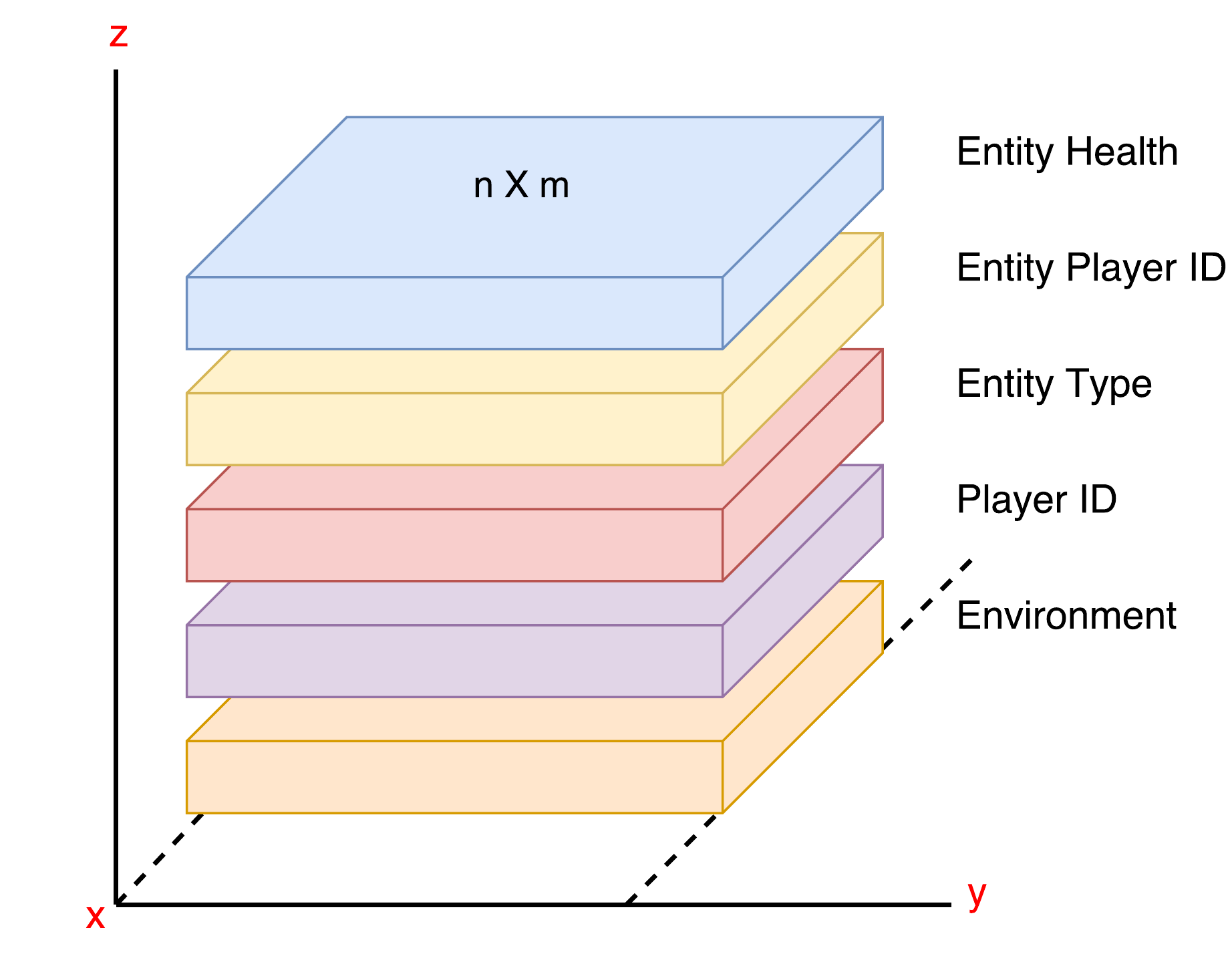}
	\caption{Illustration of how the raw state is represented using 3-D matrices}
	\label{fig:deeprts_arch}
\end{figure}

The Deep RTS game engine features two graphical interface modes in addition to the headless mode that is used by default. The primary graphical interface relies on Python while the second is implemented in C++. The Python version is not interactive and can only render the raw game-state as an image.  By using software rendering, the capture process of images is significantly faster because the copy between GPU and CPU is slow. The C++ implementation, seen in Figure \ref{fig:deep_rts_game} is fully interactive, enabling manual play of Deep RTS. Figure \ref{fig:deeprts_arch} shows how the raw game-state is represented as a 3-D matrix in headless mode. Deep learning methods often favor raw game-state data instead of image representation as sensory input. This is because raw data is often more concrete with clear patterns.

\subsection{Action-space definition}
The action-space of the Deep RTS game environment is separated into two abstract levels. The first level is actions that directly impact the environment, for instance, right-click, left-click, move-left, and select-unit. The next layer of abstraction is actions that combine actions from the previous layer, typically \textit{select-unit} $\rightarrow$ \textit{right-click} $\rightarrow$ \textit{right-click} $\rightarrow$ \textit{move-left}. The benefit of this abstraction is that algorithms can focus on specific areas within the game-state, and enable to build hierarchical models that each specialize in tasks (planning). The Deep RTS initially features 16 different actions in the first layer and 6 actions in the last abstraction layer, but it is trivial to add additional actions.

\subsection{Summary}
This section presents some of the central parts what the Deep RTS game environment features for reinforcement learning research. It is designed to measure the performance of algorithms accurately having a standardized API through OpenAI Gym, which is widely used in the reinforcement learning community.


\section{Experiments}
\label{sec:experiments}

\subsection{Performance considerations in Deep RTS}
The goal of Deep RTS is to simulate RTS scenarios with ultra high-performance accurately. The performance is measured by how fast the game engine updates the game-state, and how quickly the game-state can be represented as an image. Some experiments suggest that it is beneficial to render game graphics on the CPU instead of the GPU. Because the GPU has a separate memory, there is a severe bottleneck when copying the screen buffer from the GPU to the CPU.

\begin{figure*}[!t]
	\centering
	\subfloat[Correlation between FPS (Y-axis) and Map-sizes (X-axis)]{\includegraphics[width=2.5in]{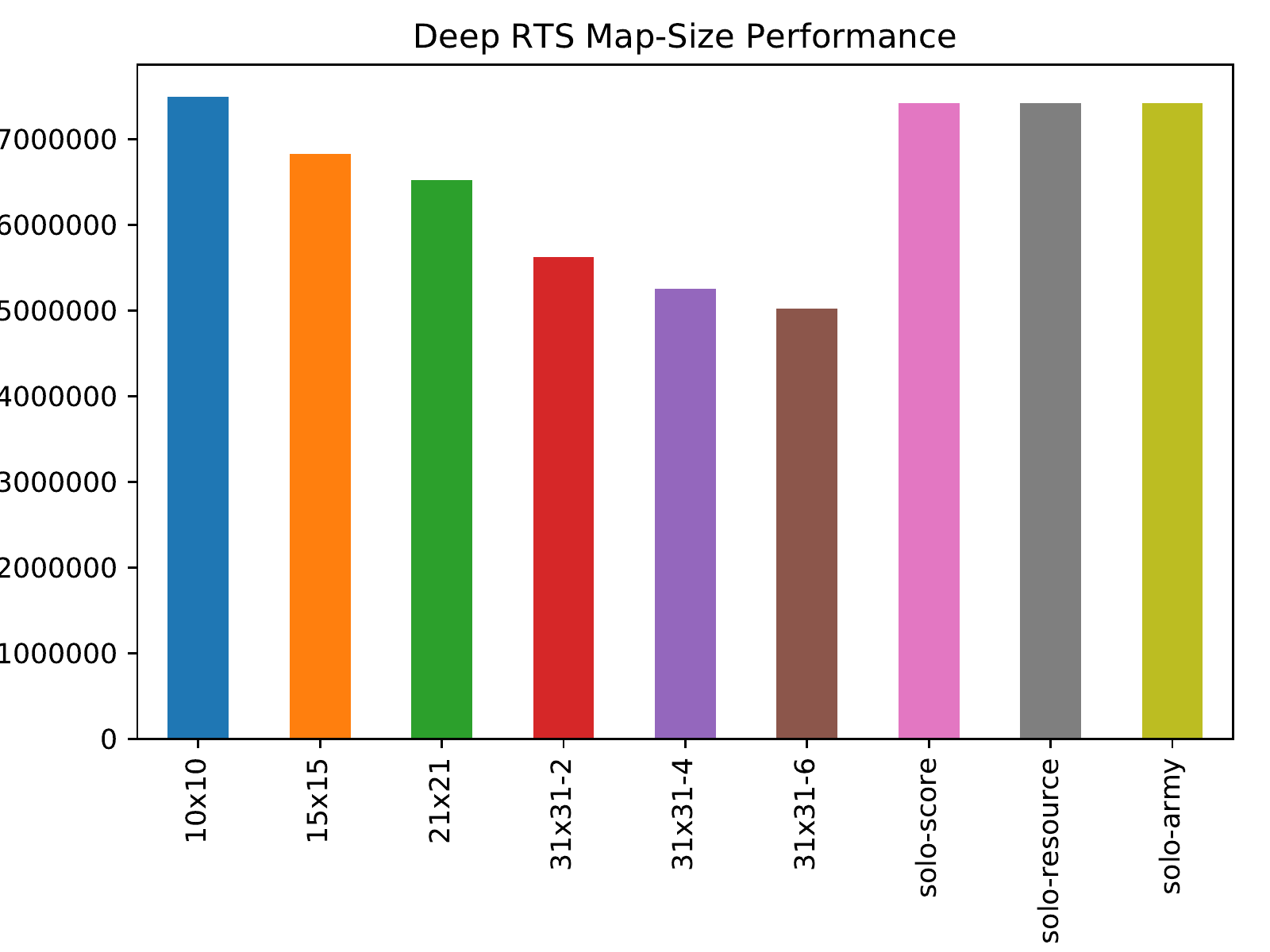}%
		\label{fig:fps_performance_map}}
	\hfil
	\subfloat[Correlation between FPS and Number of Units]{\includegraphics[width=2.5in]{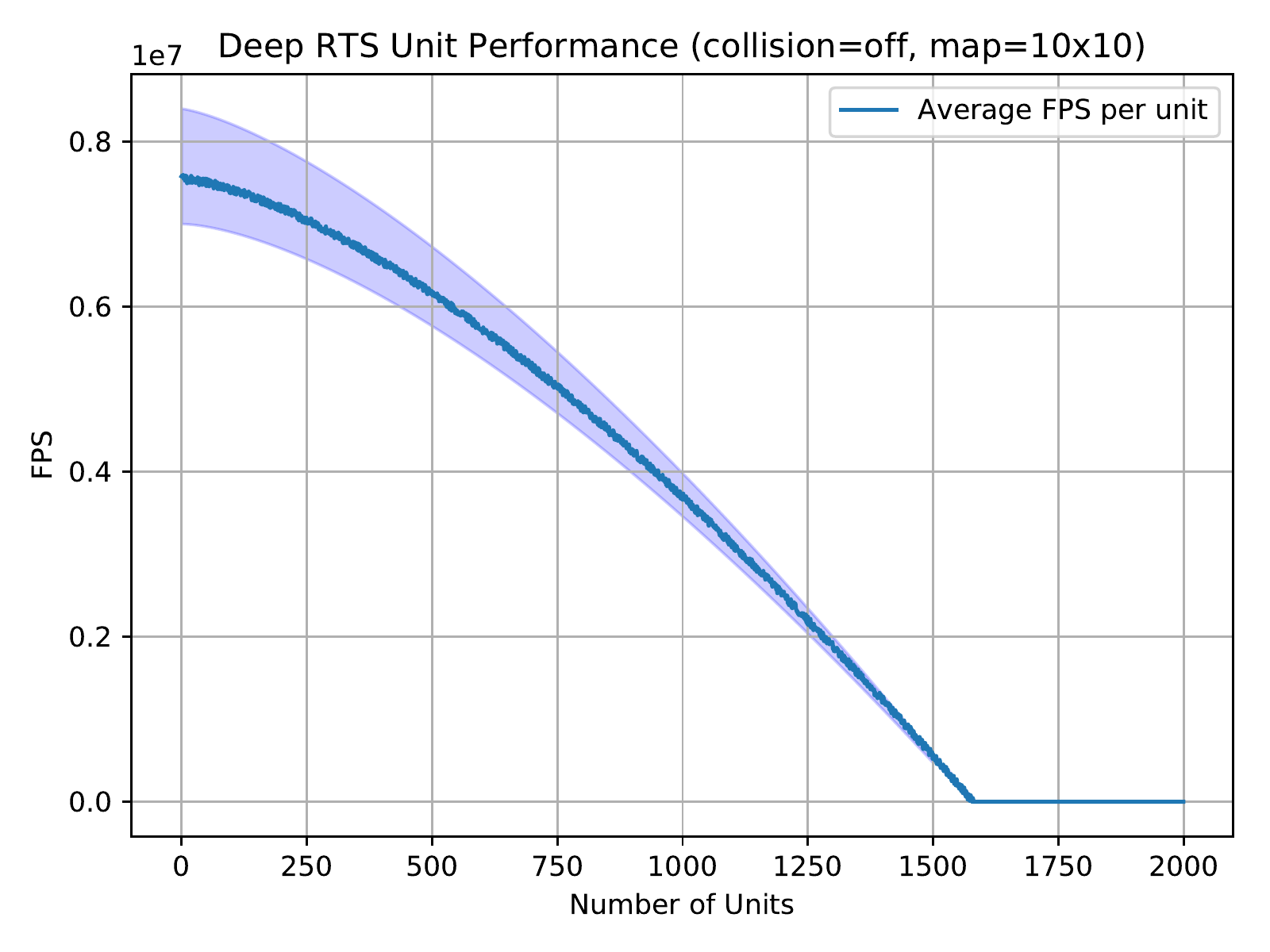}%
		\label{fig:fps_performance_num_units}}
	\caption{FPS Performance in Deep RTS}
	\label{fig:fps_performance}
\end{figure*}

Figure \ref{fig:fps_performance_map} shows the correlation between the frame-rate and size of the game map. Observing the data, it is clear that the map-size has O(n) penalty to the frame-rate performance. It is vital to preserve this linearity, and optimally have the constant performance of O(1) per game update. Figure \ref{fig:fps_performance} extends this benchmark by testing the impact a unit has on the game performance, averaging 1 000 games for all map-sizes. The data indicates that entities have an exponential impact on the frame-rate performance. The reason for this is primarily the jump-point-search algorithm used for unit path-finding. The path-finding algorithm can be disabled using custom configurations. 

The Deep RTS game environment is high-performance, with few elements that significantly reduce the frame-rate performance. While some mechanics, namely path-finding is a significant portion of the update-loop it can be deactivated by configurations to optimize the performance further.

\subsection{Comparing Deep RTS to existing learning environments}
\begin{table}[!ht]
	\centering
	\caption{Comparison of the FPS for selected environments. The Deep RTS benchmarks are performed using minimum and maximum configurations}
	\label{tbl:game_environments_fps_performance}
	\begin{tabular}{|ccc|}
		\hline
		\rowcolor{black}
		\color{white}\textbf{Environment} & \color{white}\textbf{Frame per second} & \color{white}\textbf{Source} \\ \hline
		\multicolumn{1}{|l|}{ALE} & 6,500 & \cite{Bellemare2012} \\
		\multicolumn{1}{|l|}{Malmo Platform} & ~60-144 & \cite{Johnson2016} \\
		\multicolumn{1}{|l|}{ViZDoom} & ~8,300 & \cite{Kempka2016a}\ \\
		\multicolumn{1}{|l|}{DeepMind Lab} & ~1,000 & \cite{Beattie2016} \\
		\multicolumn{1}{|l|}{OpenAI Gym} & 60 & \cite{Brockman2016} \\
		\multicolumn{1}{|l|}{OpenAI Universe} & 60 & \cite{OpenAI2016} \\
		\hline
		\multicolumn{1}{|l|}{Stratagus} & ~60-144 &  \cite{Ponsen2005 } \\
		\multicolumn{1}{|l|}{microRTS} & 11,500 &  \cite{Ontanon2013}\\
		\multicolumn{1}{|l|}{TorchCraft} & 2,500 &  \cite{Synnaeve2016} \\
		\multicolumn{1}{|l|}{ELF} & 36,000 & \cite{Tian2017} \\
		\multicolumn{1}{|l|}{SC2LE} & ~60-144 & \cite{Vinyals2017} \\
		\multicolumn{1}{|l|}{\textbf{Deep RTS}} & 24,000, 7,000,000 & - \\
		\hline
	\end{tabular}
\end{table}
There is a substantial difference between the performance in games targeted research and those aimed towards gaming. Table \ref{tbl:game_environments_fps_performance} shows that the frame-rate difference ranges from 60 to 7 000 000 for selected environments. A high frame-rate is essential because some exploration algorithms often require a quick assessment of future states through forward-search. Table \ref{tbl:game_environments_fps_performance} shows that microRTS, ELF, and Deep RTS are superior in performance compared to other game environments. Deep RTS is measured using the largest available map (Table \ref{tbl:scenario_overview}) having a unit limit of 20 per player. This yields the performance of 24 000 updates-per-second. The Deep RTS game engine can also render the game state with up to 7 000 000 updates-per-second using the minimal configuration. This is a tremendous improvement on previous work and could enable algorithms with a limited time budget to do deeper tree-searches.

\subsection{Using Deep Q-Learning in Deep RTS}
\begin{figure}[!t]
	\centering
	\includegraphics[width=\linewidth]{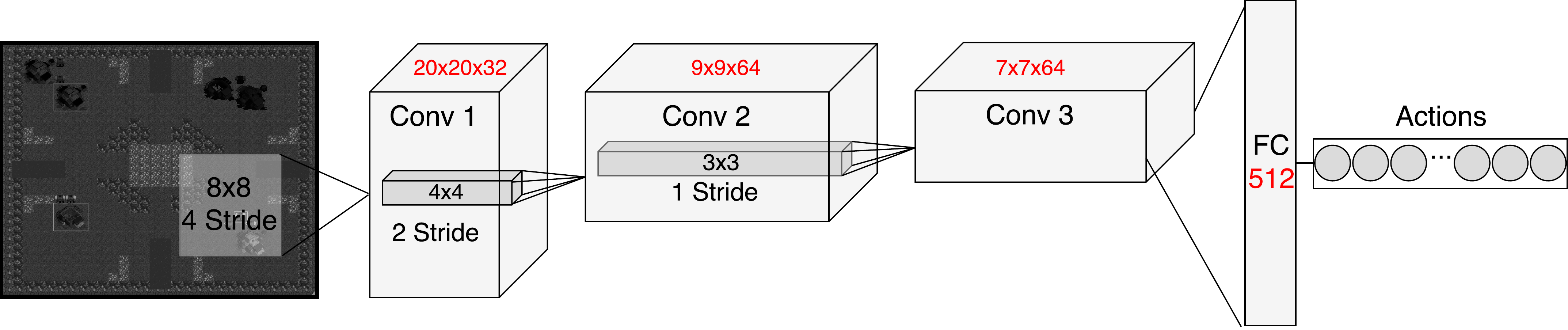}
	\caption{Overview of the Deep Q-Network architecture used in the experiments. Inspired by the work seen in \cite{Mnih2013}}
	\label{fig:convnet_arch}
\end{figure}

At the most basic level, Q-Learning utilizes a table for storing $(s,a,r,s^{'})$ pairs, where $s$ is the states, $a$ is the actions, $r$ the rewards, and $s^{'}$ the next state. Instead, a non-linear function approximation can be used to approximate $Q(s,a;\theta)$. This is called \textbf{Deep-Q Learning}. $\theta$ describes the tunable parameters (weights) for the approximation function. Artificial neural networks are used as an approximation function for the Q-Table but at the cost of stability~\cite{Mnih2015}. Using artificial neural networks is much like compression found in JPEG images. The compression is \textit{lossy}, and some information is lost during the compression. Deep Q-Learning is thus unstable, since values may be incorrectly encoded during training~\cite{Lillicrap2015}.

\begin{figure}[!t]
	\centering
	\includegraphics[width=2.80in]{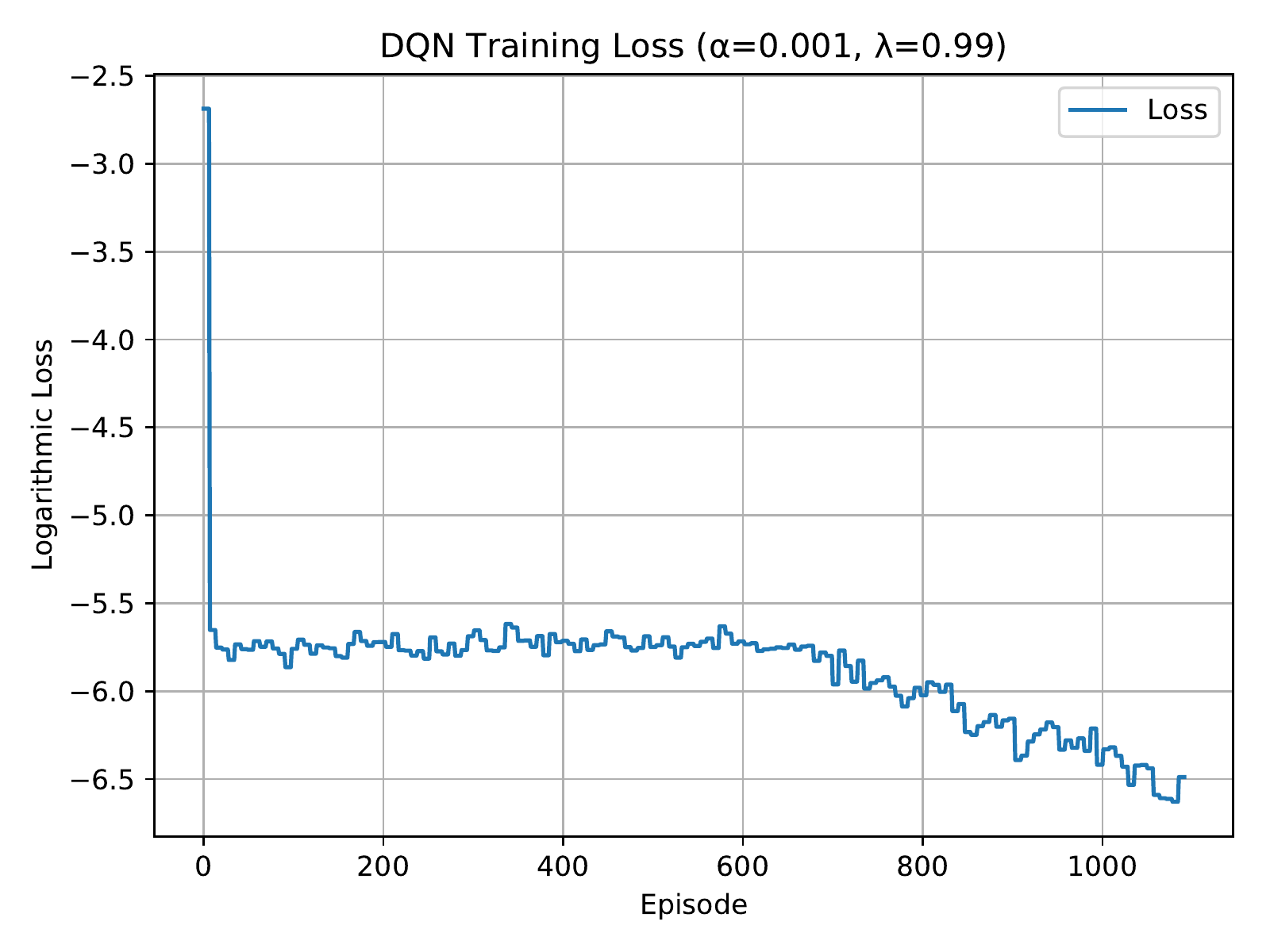}
	\caption{Training loss of the Deep Q-Network. Each episode consists of approximately 1 000 epochs.}
	\label{fig:training_loss_dqn}
\end{figure}

This paper presents experimental results using the Deep Q-Learning architecture from \cite{Mnih2015, VanHasselt2015}. Figure~\ref{fig:convnet_arch} shows the network model, and figure~\ref{fig:training_loss_dqn} illustrates the averaged training loss of 100 agents. The agent uses gray-scale image game-state representations with an additional convolutional layer to decrease the training time, but can also achieve comparable results after approximately 800 episodes of training with the exact architecture from~\cite{Mnih2015}\footnote{Each episode contains approximately 1 000 epochs of training with a batch size of 16}. The graph shows that the agent quickly learns the correlation between game-state, action and the reward function. The loss quickly stabilizes at a relatively low value, but it is likely that very small optimizations in the parameters have a significant impact on the agent's performance.

\begin{figure*}[!t]
	\centering
	\subfloat[DQN vs Random-play AI in the 15x15-2-FFA scenario]{\includegraphics[width=2.5in]{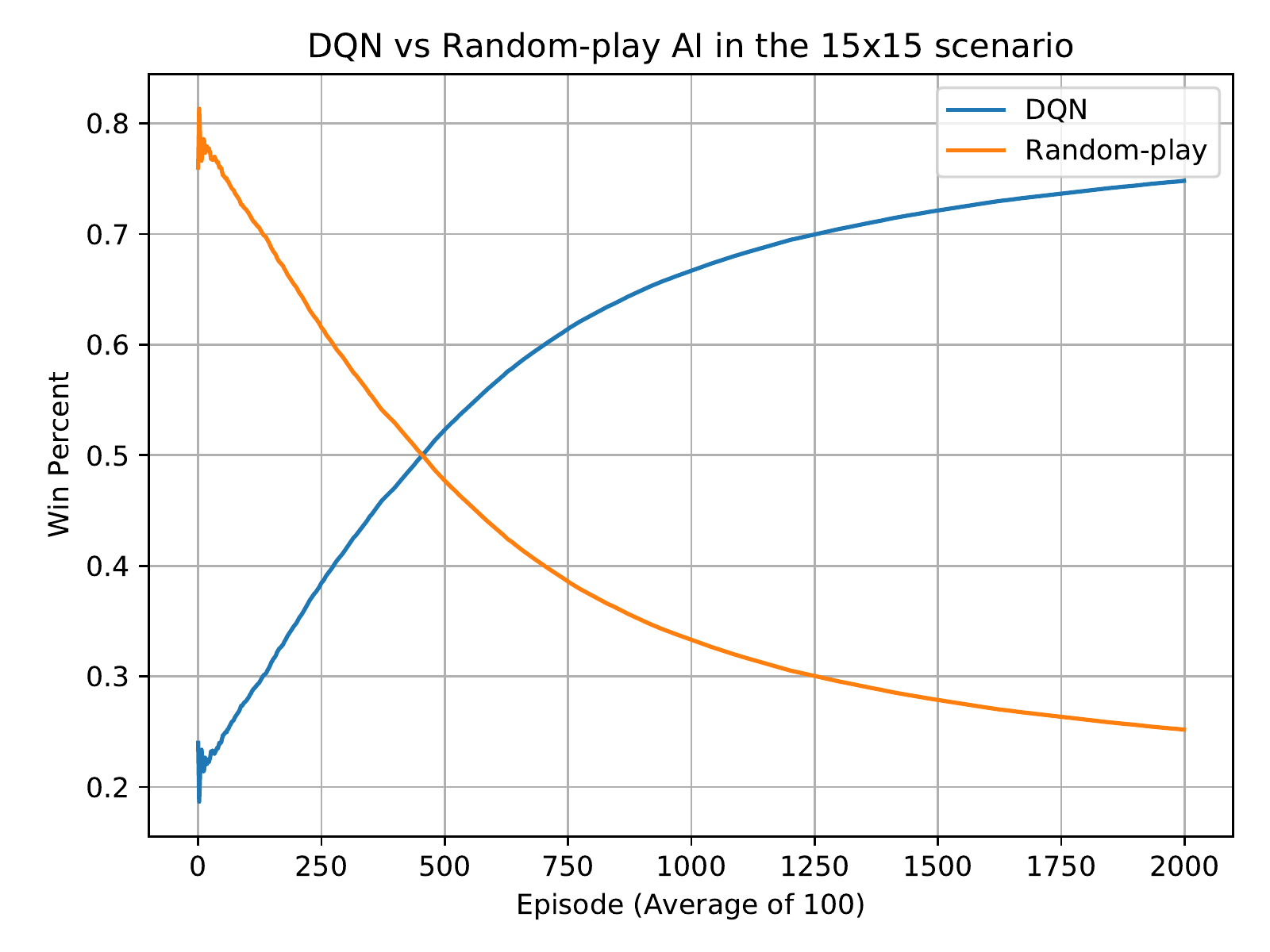}%
		\label{fig:agent_performance_random}}
	\hfil
	\subfloat[DQN vs Rule-based AI in the 15x15-2-FFA scenario]{\includegraphics[width=2.5in]{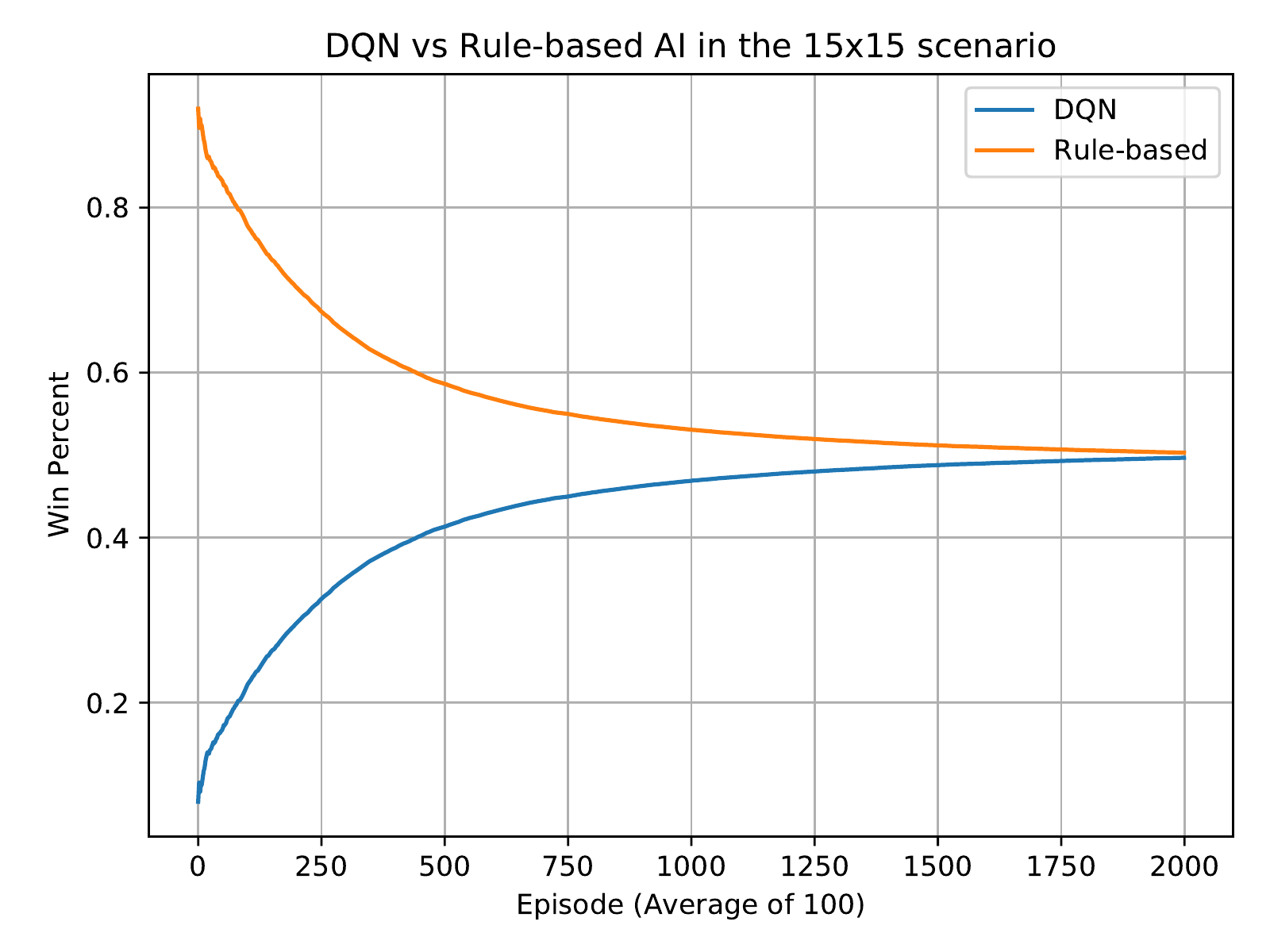}%
		\label{fig:agent_performance_rule_based}}
	\caption{Performance comparison of agents using  Deep Q-Network, random-play, and rule-based strategies}
	\label{fig:dqn_rule_based_random}
\end{figure*}

Figure~\ref{fig:agent_performance_random} shows the win-rate against an AI with a random-play strategy. The agent quickly learns how to perform better than random behavior, and achieves 70~\% win-rate at episode 1 250. Figure~\ref{fig:agent_performance_rule_based} illustrates the same agent playing against a rule-based strategy. The graph shows that the Deep Q-Network can achieve an average of 50~\% win-rate over a 1 000 games. This strategy is considered an easy to moderate player, where its strategy is to expand the base towards the opponent and build a military force after approximately 600 seconds. Figure~\ref{fig:deep_rts_game} shows how the rule-based player (blue) expands the base to gain the upper hand.

The experimental results presented in this paper show that the Deep RTS game environment can be used to train deep reinforcement learning algorithms. The Deep Q-Network does not achieve super-human expertise but performs similarly to a player of easy to moderate skill level, which is a good step towards a high-level AI.

\section{Conclusion and Future Work}
\label{sec:conclusion}
This paper is a contribution towards the continuation of research into deep reinforcement learning for RTS games. The paper summarizes previous work and outlines the few but essential success stories in reinforcement learning. The Deep RTS game environment is a high-performance RTS simulator that enables rapid research and testing of novel reinforcement learning techniques. It successfully fills the gap between the vital game simulator microRTS, and StarCraft II, which is the ultimate goal for reinforcement learning research for the RTS game genre.

The hope is that Deep RTS can bring insightful results to the complex problems of RTS~\cite{Ontanon2013} and that it can be a useful tool in future research.

Although the Deep RTS game environment is ready for use, several improvements can be applied to the environment. The following items are scheduled for implementation in the continuation of Deep RTS:
\begin{itemize}
	\item Enable LUA developers to use Deep RTS through LUA bindings.
	\item Implement a generic interface for custom graphics rendering.
	\item Implement duplex WebSockets and ZeroMQ to enable any language to interact with Deep RTS
	\item Implement alternative path-finding algorithms to increase performance for some scenarios
	\item Add possibility for memory-based fog-of-war to better mimic StarCraft II
\end{itemize}

\section{Acknowledgements}
We would like to thank Santiago Ontanon at the University of Drexel for his excellent work on microRTS. microRTS has to us been  a valuable tool for benchmarks in the reinforcement learning domain and continues to be the goto environment for research in tree-search algorithms.

\bibliographystyle{IEEEtran}
%

\end{document}